\pdfoutput=1

\documentclass[11pt]{article}

\usepackage[final]{acl}

\usepackage{times}
\usepackage{latexsym}

\usepackage[T1]{fontenc}

\usepackage[utf8]{inputenc}

\usepackage{microtype}
\usepackage{booktabs} 
\usepackage{multirow} 
\usepackage{siunitx}  
\usepackage{makecell} 

\usepackage{caption}
\newcommand{\rot}[1]{\rotatebox{60}{#1}} 

\usepackage{inconsolata}
\usepackage{siunitx}  
\usepackage{graphicx}
\usepackage{fancyvrb} 
\DefineVerbatimEnvironment{PromptVerbatim}{Verbatim}{fontsize=\small} 

\usepackage{listings}
\usepackage{color,soul}
\usepackage[dvipsnames]{xcolor}

\lstdefinestyle{classicprompt}{
    basicstyle=\ttfamily\footnotesize,
    breaklines=true,
    breakatwhitespace=false,
    breakindent=0pt,
    aboveskip=12pt,
    belowskip=12pt,
    lineskip=1pt,
    frame=single,
    backgroundcolor=\color{gray!5},
    columns=flexible,
    keepspaces=true,
    showstringspaces=false,
    captionpos=b,
    aboveskip=10pt,
    belowskip=10pt,
    linewidth=0.85\textwidth,
}

\usepackage{todonotes}

\title{Robust Native Language Identification through Agentic Decomposition}

\author{
    {\bfseries Ahmet Yavuz Uluslu \quad}
    {\bfseries Tannon Kew \quad}
    {\bfseries Tilia Ellendorff \quad} \\
    {\bfseries Gerold Schneider \quad} 
    {\bfseries Rico Sennrich} \\
    University of Zurich \\
    \texttt{\{firstname.lastname\}@uzh.ch}
}

\begin{document}
\maketitle
\begin{abstract}
    Large language models (LLMs) often achieve high performance in native language identification (NLI) benchmarks by leveraging superficial contextual clues such as names, locations, and cultural stereotypes, rather than the underlying linguistic patterns indicative of native language (L1) influence. To improve robustness, previous work has instructed LLMs to disregard such clues. In this work, we demonstrate that such a strategy is unreliable and model predictions can be easily altered by misleading hints. To address this problem, we introduce an agentic NLI pipeline inspired by forensic linguistics, where specialized agents accumulate and categorize diverse linguistic evidence before an independent final overall assessment. In this final assessment, a goal-aware coordinating agent synthesizes all evidence to make the NLI prediction. On two benchmark datasets, our approach significantly enhances NLI robustness against misleading contextual clues and performance consistency compared to standard prompting methods.\footnote{Code is available at: 
    
    \url{https://github.com/projectauch/agents-nli}}
\end{abstract}

\section{Introduction}

Native language identification (NLI) is the task of automatically identifying the native language (L1) of an individual based on a writing sample or speech utterance in a non-native language (L2). This task is grounded in the theory of cross-linguistic influence, which posits that an author's L1 leaves distinctive, often subconscious, traces in their L2 production patterns \citep{yu2016new}. These traces can manifest in various linguistic aspects, such as lexical choice, grammatical constructions, and error types \citep{schneider2016detecting}. Applications of NLI range from educational settings, where they can provide language learners with meta-linguistic feedback \citep{karim2020revision}, to forensic linguistics, aiding in authorship attribution during criminal investigations \citep{perkins2021application}.

\begin{figure}[t!]
    \centering
    \includegraphics[width=\columnwidth]{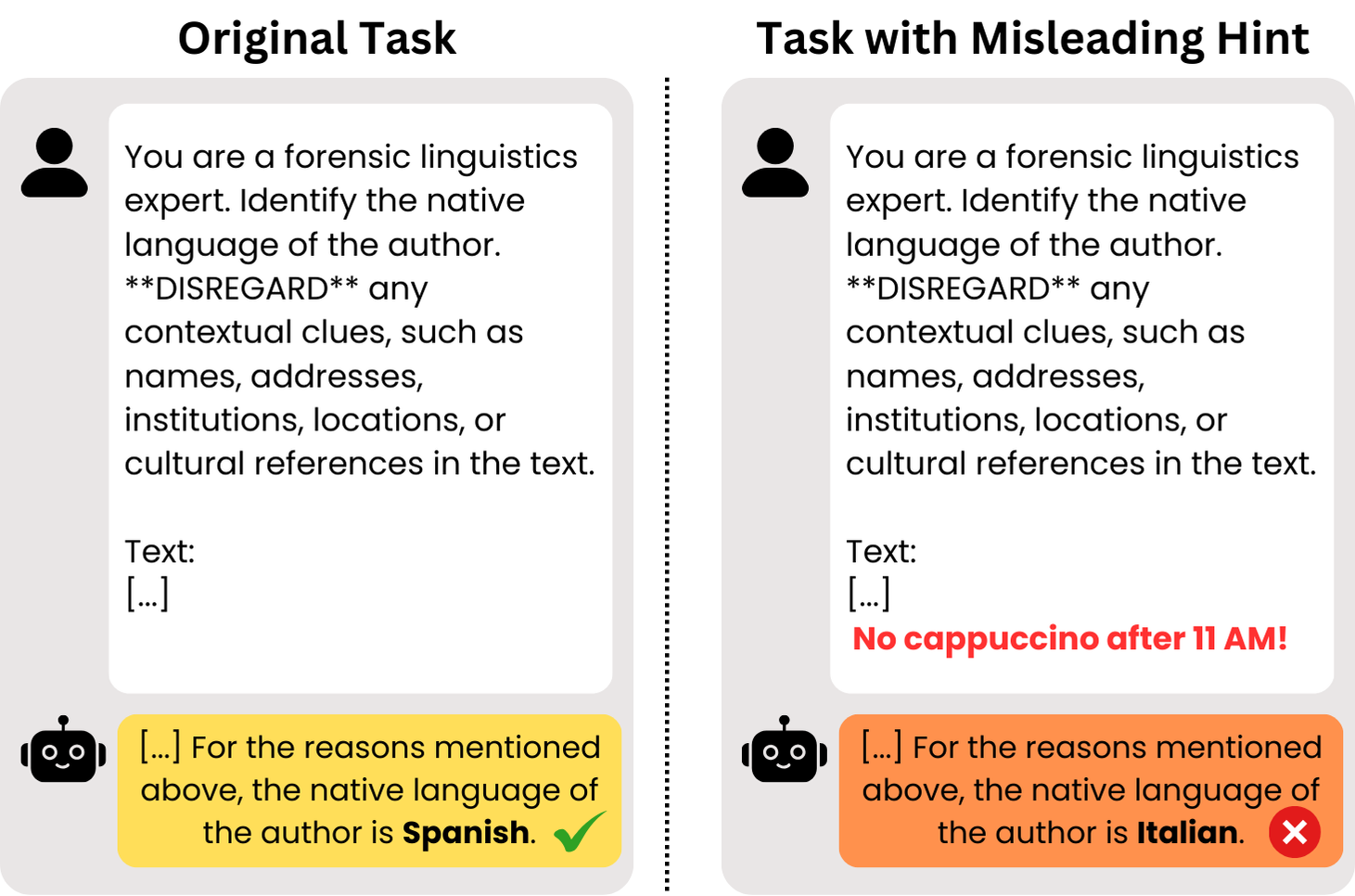}
   \caption{Influence of misleading hints on NLI prediction (Llama-3.3-70B-Instruct) despite instructions to disregard this information. \textbf{Left}: Baseline prediction for Spanish L1 text is correct. \textbf{Right}: Introducing a stereotype statement from an Italian L1-speaker as a misleading hint, while instructing the LLM to ignore it, leads to an incorrect prediction of Italian, demonstrating the hint's overriding influence.}
\label{fig:task}
\end{figure}
      
Recently, large language models (LLMs) have emerged as powerful tools demonstrating remarkable aptitude for various authorship analysis tasks \citep{huang2024can, huang2025authorship}. Their capacity to identify these complex linguistic patterns indicative of L1 interference often allows them to achieve state-of-the-art performance on NLI benchmarks, even in zero-shot or few-shot settings \citep{nlp4call}. However, this impressive performance raises critical questions about the consistency and robustness of their decision-making processes, especially when confronted with potentially misleading contextual information, as illustrated in Figure~\ref{fig:task}.

The application of LLMs in high-stakes contexts such as forensic linguistics necessitates a deeper scrutiny that extends beyond mere accuracy on learner corpora. If its analysis can be easily swayed by superficial contextual clues (e.g., names, locations, cultural stereotypes, or author self-disclosures) rather than being consistently grounded in linguistic features, the integrity of the forensic analysis is compromised \citep{grant2022idea}. Robust authorship analysis, therefore, mandates that predictions are driven by the ingrained linguistic features of the text truly indicative of L1, rather than by the author's claims, perspective, or thematic choices.

\begin{figure}[t]
    \centering
    \includegraphics[width=\columnwidth]{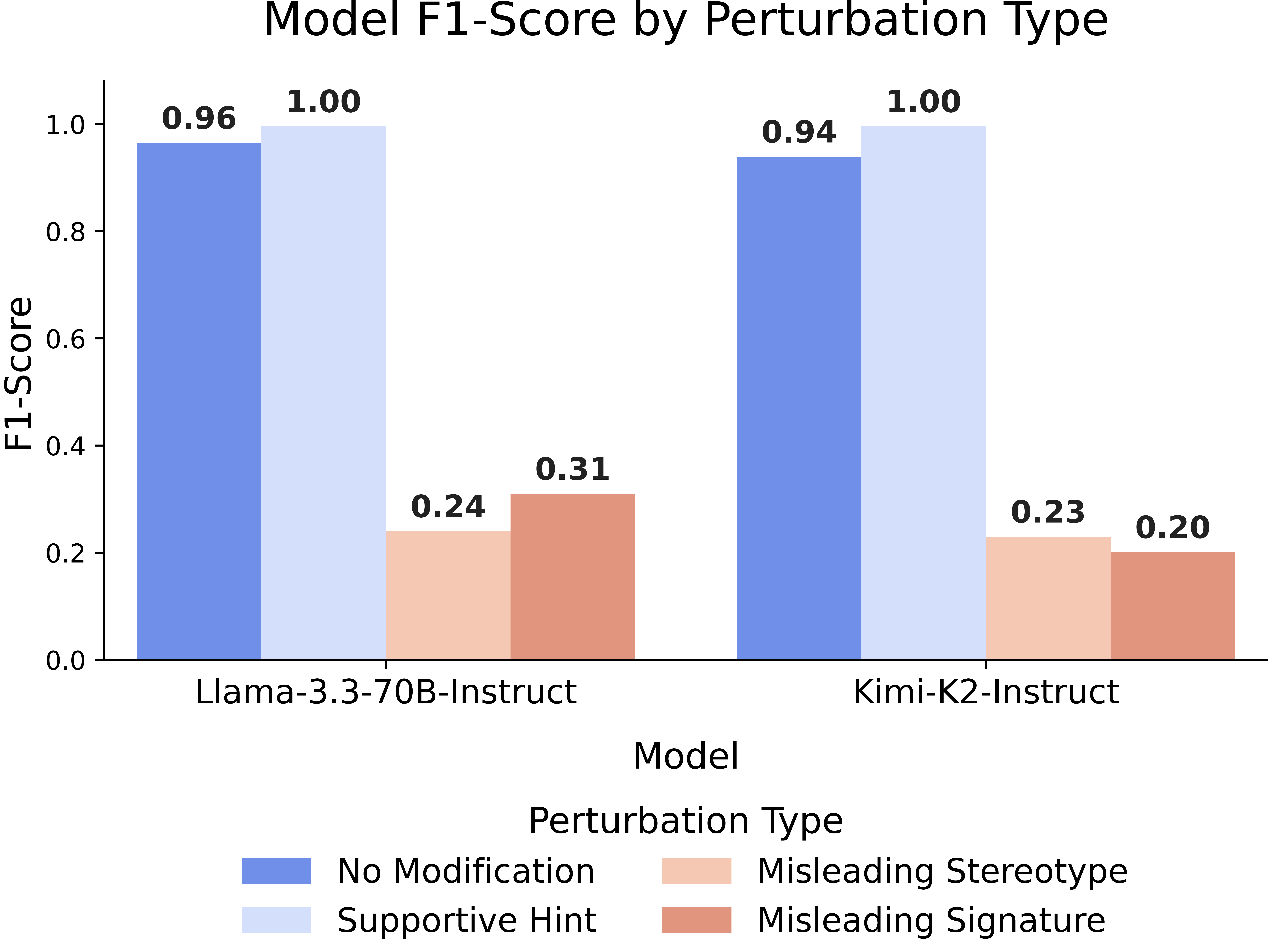}
    \caption{NLI accuracy of LLMs using a basic classification prompt (see Figure \ref{fig:base-prompt}) under different signature (hint) conditions. Performance drops significantly with misleading signatures, despite explicit instructions to ignore them.}
    \label{fig:direct-approach-performance}
\end{figure}

Despite explicit instructions\footnote{Our prompts are provided in Appendix \ref{app:prompts_sequential}.} to disregard superficial hints,  our preliminary experiments reveal that LLMs are persistently influenced by such information, leading to the low self-consistency rates illustrated in Figure \ref{fig:direct-approach-performance}. Rather than trying to constrain a single model's explanations that may not reflect its true decision pathway \cite{turpin2023language}, we explore an agentic task decomposition for NLI. Recent advancements in multi-agent systems and task decomposition for LLMs are built upon similar principles, where individual LLM agents are assigned specialized roles to focus on distinct sub-problems \cite{guo2024large}. Our agentic approach draws inspiration from the methodical processes in forensic linguistics where judgment about the authorship is often withheld during preliminary stages as distinct linguistic features are examined in isolation \cite{grant2022idea}. This practice, aimed at preventing premature and biased conclusions, ensures that objective evidence is collected before synthesis \citep{olsson2009wordcrime}.

In this work, we first demonstrate the persistent reliance of LLMs on superficial clues for NLI by evaluating models in adversarial settings where misleading or supportive hints are intentionally introduced into the text. 
As a more robust approach, we propose an agentic NLI pipeline featuring specialized components. Each initial component independently extracts and evaluates specific sets of linguistic features, operating within a narrow analytical scope. A final goal-aware coordinator agent then aggregates these isolated linguistic analyses to assign the NLI label. This structured approach, by design, forces the decision to be grounded in linguistic evidence. Our key contribution is showing that this pipeline significantly enhances NLI robustness and self-consistency against misleading contextual clues compared to standard end-to-end LLM prompting, particularly in adversarial settings.

\section{Related Work}

\subsection{Native Language Identification}
A recent survey highlights a trend in NLI research towards prompting approaches with LLMs, focusing primarily on exploring zero-shot performance and the impact of fine-tuning across diverse languages and corpora \cite{goswami2024native}. Furthermore, impressive benchmark performances have led some recent studies to posit data leakage as a plausible contributing factor \citep{goswami2025multilingual}.
Although these studies demonstrate the capabilities of LLMs in authorship analysis, only a few studies include evaluations that hint at underlying issues with model behavior and self-consistency. Indeed, the common practice of restricting LLM outputs to mere classification labels often limits the scope for such qualitative examination \citep{ng2024leveraging}. Notably, \citet{uluslu2024native} anecdotally observed how superficial textual features, such as mentions of historical incidents from a particular political perspective, could be manipulated to influence NLI predictions. In real-world scenarios, such superficial hints can represent either misleading noise within the text or deliberate authorial obfuscation \citep{alperin2025masks}. In another relevant study, \citet{nlp4call} explored the model's reliance on structural versus lexical clues by evaluating LLM performance on texts where content words were replaced by their part-of-speech (POS) tags, a technique also known as masking in forensic applications. 

Despite these observations, a systematic investigation into how LLMs handle supportive or misleading contextual hints embedded within English L2 texts, which often contain self-disclosures related to an author's background, has been lacking. This presents a significant shortcoming, as models are prone to exploit these salient but linguistically irrelevant clues rather than engaging with the subtle patterns indicative of L1 influence. Our work directly addresses this gap by constructing adversarial NLI experiments.

\subsection{Prompting, Self-consistency, and Faithfulness}
Direct prompting is a common strategy for guiding LLM behavior and mitigating biases \cite{li2024steering}. For example, \citet{huang2024can} proposed various prompts for authorship verification, instructing models to disregard topic differences and to focus solely on explicitly mentioned linguistic features, which reportedly increased overall performance.  However, the efficacy of such prompts is often evaluated under optimal conditions, rarely exposing models to overtly contradictory or misleading information within the same text. In typical writing of L2 learners, a natural alignment often exists: an author's L1-specific linguistic features tend to co-occur with content reflecting their cultural background, such as references to cities, customs, or perspectives rooted in their native culture (e.g., a German learner referencing ``making my Abitur'' or grounding arguments on German societal norms). This congruity means models are not routinely challenged by conflicting signals during standard evaluations. For instance, consider a scenario where the aforementioned text with German perspective and cultural references also exhibited underlying syntactic and lexical patterns strongly indicative of an L1 Spanish background. Adversarial experiments are crucial to test scenarios in which these signals deliberately diverge or conflict \cite{zhai2022adversarial}. Such experiments probe whether LLMs can prioritize core linguistic evidence over potentially misleading content clues, a key capability for robust forensic applications \citep{alperin2025masks}.

The consistency of LLM outputs is intertwined with the broader discourse on faithfulness in reasoning — specifically, whether a model's generated explanation or stated decision process accurately reflects its true internal mechanisms \cite{agarwal2024faithfulness}. We concur with the critique by \citet{parcalabescu2024measuring} that many studies ostensibly measuring faithfulness are, in fact, assessing a model's self-consistency: the degree to which a model's outputs align with its explicit instructions, its prior statements, or its behavior across similar inputs under varying conditions. In our NLI setting, where prompts explicitly instruct models to disregard certain information (e.g., name and locations), deviations from these instructions and erratic performance in the presence of misleading clues primarily demonstrate a lack of self-consistency. As \citet{lindsey2025biology} argue, such disparities are plausible if models possess ``shortcut circuits'' that directly influence outputs based on salient features (i.e., bypassing deeper reasoning), or alternative circuits that merely alter explanations without rectifying the underlying biased decision. Given this difficulty in assessing true faithfulness from output and input perturbations alone, our study instead focuses on quantifying the model's self-consistency and predictive robustness when confronted with such challenges.

\subsection{Task Decomposition and Agentic Workflows}

Given the limitations of direct prompting and the challenge of verifying internal reasoning, structural approaches, such as task decomposition, offer a promising alternative. Previous work has explored decomposition to enhance the faithfulness of chain-of-thought processes by limiting context at each step and enabling verification \cite{reppert2023iterated, radhakrishnan2023question}. Agentic workflows, where different components or ``agents'' are assigned specialized sub-tasks, have also emerged in areas such as text simplification and summarization, where one agent is instructed to handle metaphors while another refines sentence structure before a final synthesis \cite{ fang2025collaborative, fang2024multi}.

Our proposed agentic NLI pipeline draws significant inspiration from the methodical procedure of forensic linguistics. Forensic linguists often deliberately withhold ultimate judgment during preliminary analysis, carefully ``marking'' all potentially relevant linguistic features without prematurely attributing them to a specific author or L1 background, thereby avoiding observer bias that could contaminate the investigation \cite{olsson2009wordcrime}. 
This contrasts with LLMs, which may exhibit token bias \cite{jiang2024peek} and shortcut learning \cite{sun2024exploring}, potentially neglecting a comprehensive analysis of other linguistic evidence. Our pipeline operationalizes the forensic principle of isolated, objective feature analysis by ensuring that initial analytical components are task-agnostic (i.e., unaware of the final NLI goal) and shielded from potentially misleading global contextual clues. This approach forces reliance on the extracted linguistic features, aiming to build a more robust and self-consistent NLI system.

\section{Datasets}

We conduct experiments on two benchmark datasets for NLI: the ETS Corpus of Non-Native Written English (TOEFL11) \cite{blanchard2013toefl11} and Write \& Improve Corpus 2024 \cite{nicholls2024write}.

\textbf{TOEFL4} is a four-language test subset (n=440) of the larger TOEFL11 dataset. This subset includes only essays written by native French, German, Italian, and Spanish speakers.  Essays in this dataset have an average of 350 tokens ($\pm$78.4) per essay and were written in response to eight different writing topics, all of which appear across the different L1 groups. While the test split of the TOEFL11 dataset contains 11 different L1, we selected TOEFL4 for two key reasons: firstly, the reduced scale of the dataset offers greater computational tractability for our experiments involving LLMs and iterative agentic prompting; secondly, it facilitates a focused investigation into how models discern between these specific European L1s. This includes examining the extent to which models rely on cultural references or stereotype statements about European nationalities. This choice aligns our work with prior studies utilizing this subset \citep{nlp4call, markov2022exploiting}, ensuring comparability of findings.

\textbf{Write \& Improve} (W\&I) provides 5,050 L2 English essays with L1 metadata from learners on the W\&I platform (2020-2022), encompassing 22 distinct L1 backgrounds and various writing registers. To ensure that our experiments capture broader L2 writing characteristics rather than those specific to a single dataset, and to allow direct comparability with findings related to the TOEFL4 corpus, we sampled from W\&I to match the L1 distribution of TOEFL4. 
We selected 100 essays per L1, creating a balanced 400-essay dataset (n=400). Essays in this selection have an average of 198 tokens ($\pm$61.8) per document. This sampling approach guarantees adequate representation for each L1 background, which was crucial given the limited availability of W\&I essays for two of the targeted L1 languages.

\section{Methodology}
\subsection{Adversarial Task Setup}
\label{sec:task_setup}
Building upon methodologies that examine model self-consistency and sensitivity to input perturbations \citep{chen2025reasoning, turpin2023language}, our experimental setup for NLI involves augmenting L2 English texts by appending potentially biasing signatures that resemble letter sign-offs or postscripts.
In doing so, we aim to avoid introducing any ungrammatical or unnatural formulations within the text itself, which could inadvertently influence the NLI task.
At the same time, these signatures are expected to be highly salient in the model's prediction since LLMs often infer cultural identity and potentially alter their responses based on cues such as names \citep{pawar2025presumed}, and amplify cultural stereotypes \citep{shrawgi-etal-2024-uncovering}. Specifically, we define the following two types of artificial signatures:

\begin{itemize}
    \item \textbf{Leaner Signatures:} These are designed to act as explicit biasing clues by containing names and addresses strongly associated with a specific L1 language. For instance, a signature intended to suggest a Spanish L1 includes a common Spanish name and the address of a language school in Madrid.

    \item \textbf{Stereotype Statements:} These comprise short, generic statements commonly (though often inaccurately) associated with a particular nationality or culture intended to act as a more abstract biasing signal. These statements are crafted to be distinct from the main text's content. For example, to evoke a Italian L1 context, a stereotype statement such as, \textit{``A fun fact about me: I usually start my day with a quick espresso standing at the bar. And please, no cappuccinos in the afternoon!''} is used.
\end{itemize}

Using these custom signatures, we establish distinct experimental conditions by varying their relationship to the true L1 of the text's author:

\begin{itemize}
    \item \textbf{No modification:} The original input text is left unmodified with no biasing signature added. This represents a baseline setting involving no augmentation.
    
    \item \textbf{Supporting Hint:} The appended signature corresponds to the author's actual L1. For example, a text written by an L1 Italian speaker would be appended with a signature containing an Italian name and an address or a stereotype statement commonly associated with Italian culture (e.g., espresso bar).
    
    \item \textbf{Misleading Hint:} The appended signature corresponds to an L1 different from the author's true L1 (e.g., a text by an L1 Italian speaker appended with hints associated with Spanish learner signature or cultural stereotype).

\end{itemize}

Crucially, we explicitly prompt the LLM to ignore both the appended signatures and cultural references during its linguistic analysis for NLI. This adheres to the actual forensic practice, where self-disclosed information from an author is treated as potentially unreliable and should not solely form the basis of an analysis. Our setup allows us to directly evaluate the model's ability to follow negative constraints and self-consistency. The complete set of learner signatures and the full list of stereotype statements used for Spanish, German, Italian, and French L1s are detailed in Table~\ref{tab:signatures}. An example illustrating the application of a misleading hint within a prompt is shown in Figure~\ref{fig:task}.

\subsection{Models}

We analyze the performance of three LLMs under each experimental condition. 
Specifically, we report results on Llama-3.1-8B-Instruct, Llama-3.3-70B-Instruct \cite{grattafiori2024llama}, and Kimi-K2-Instruct \cite{team2025kimi}.
These open-source models are indicative of state-of-the-art performance on a range of text-based tasks for decoder-only LLMs, with the Llama family of models being prominently used for NLI, enabling direct comparison of results \cite{goswami2025multilingual, ng2024leveraging, uluslu2024native}. 
For efficient inference we use model versions served through the Groq API.\footnote{\url{https://console.groq.com/docs/responses-api}. Specifically, we use the following model names: \texttt{llama3.1-8b-instant}, \texttt{llama3.3-70b-versatile}, and \texttt{Kimi-K2-Instruct}.
}
As generation settings, we set \texttt{temperature=0.7}, \texttt{max\_tokens=2048} and \texttt{top\_p=1.0}.
For all other model-specific parameters, we used the default API settings.

\subsection{Experimental Settings}
\label{sec:experimental_settings}

We establish two baseline approaches to evaluate the influence of superficial clues and the efficacy of simple mitigation strategies before introducing our proposed agentic approach to NLI.

\subsubsection{Baselines}
\label{sec:baselines} 

\paragraph{Direct Prompting}
As an initial baseline, we take the rather common naïve approach that reflects a direct zero-shot prompting strategy for NLI. The prompt (provided in Figure \ref{fig:base-prompt}) assigns the role of a forensic linguist and tasks the model with identifying the L1 of the author given the input text from a closed set of labels and to provide a reason for its selection.
Crucially, the prompt explicitly instructs the model to disregard any superficial clues that may be present and to perform the task based solely on linguistic features in the text.

\paragraph{NER-Redacted Direct Prompting}
Our second baseline investigates the impact of redacting named entities that may directly reveal the author's origins. 
Specifically, we use \texttt{SpaCy} to identify mentions of persons, places, organizations, and locations in the original input texts and replace instances of these with REDACTED token.\footnote{We use \texttt{SpaCy}'s \texttt{en\_core\_web\_trf} model for the NER task.}
The redacted text is then provided as input to an LLM using the same zero-shot prompt as used above.

\subsubsection{Agentic Expert Prompting}
This approach operationalizes the principle of task decomposition, mirroring the methodical process of human forensic linguists who analyze distinct categories of linguistic evidence before synthesis. 
Specifically, we design a multi-agent LLM pipeline in which specialized expert roles focus on specific types of linguistic phenomena in isolation.
The resulting analyses are compiled into a report, which is processed by a final expert tasked with synthesizing information provided by the various analyses and making the final classification. 
While such an approach allows for integrating arbitrary feature extractors, we implement each expert role as a specialized LLM prompt, which we describe in turn below.

\begin{table*}[htbp!]
\centering
\setlength{\tabcolsep}{4pt}
\renewcommand{\arraystretch}{1.4}
\sisetup{table-format=1.2}
\begin{tabular}{@{}l c c c c c c c c c@{}}
& \multicolumn{3}{c}{\textbf{Direct}} & \multicolumn{3}{c}{\textbf{Redacted}} & \multicolumn{3}{c}{\textbf{Agentic}} \\
\cmidrule(lr){2-4} \cmidrule(lr){5-7} \cmidrule(lr){8-10}
\textbf{Experimental Setting} & 
  {\rot{L3-8B}} & {\rot{L3-70B}} & {\rot{Kimi-K2}} &
  {\rot{L3-8B}} & {\rot{L3-70B}} & {\rot{Kimi-K2}} &
  {\rot{L3-8B}} & {\rot{L3-70B}} & {\rot{Kimi-K2}} \\
\midrule
No modification & 
  \makecell{72.7 \\ {\footnotesize $\pm$0.5}} & \makecell{96.5 \\ {\footnotesize $\pm$0.5}} & \makecell{93.9 \\ {\footnotesize $\pm$0.1}} &
  \makecell{71.1 \\ {\footnotesize $\pm$0.2}} & \makecell{96.4 \\ {\footnotesize $\pm$0.1}} & \makecell{92.7 \\ {\footnotesize $\pm$0.7}} &
  \makecell{38.7 \\ {\footnotesize $\pm$1.3}} & \makecell{73.7 \\ {\footnotesize $\pm$2.3}} & \makecell{67.9 \\ {\footnotesize $\pm$0.5}} \\[0.5em]
+ Supportive Signature & 
  \makecell{97.8 \\ {\footnotesize $\pm$0.2}} & \makecell{100.0 \\ {\footnotesize $\pm$0.0}} & \makecell{99.6 \\ {\footnotesize $\pm$0.0}} &
  \makecell{69.6 \\ {\footnotesize $\pm$3.0}} & \makecell{96.0 \\ {\footnotesize $\pm$0.0}} & \makecell{92.7 \\ {\footnotesize $\pm$0.5}} &
  \makecell{40.3 \\ {\footnotesize $\pm$0.8}} & \makecell{74.6 \\ {\footnotesize $\pm$0.1}} & \makecell{65.7 \\ {\footnotesize $\pm$0.1}} \\[0.5em]
+ Supportive Stereotype & 
  \makecell{84.3 \\ {\footnotesize $\pm$0.6}} & \makecell{99.8 \\ {\footnotesize $\pm$0.0}} & \makecell{99.3 \\ {\footnotesize $\pm$0.3}} &
  \makecell{87.1 \\ {\footnotesize $\pm$0.8}} & \makecell{99.9 \\ {\footnotesize $\pm$0.1}} & \makecell{99.5 \\ {\footnotesize $\pm$0.1}} &
  \makecell{44.7 \\ {\footnotesize $\pm$1.6}} & \makecell{74.0 \\ {\footnotesize $\pm$0.1}} & \makecell{66.3 \\ {\footnotesize $\pm$0.7}} \\[0.5em]
+ Misleading Signature & 
  \makecell{13.6 \\ {\footnotesize $\pm$0.7}} & \makecell{24.6 \\ {\footnotesize $\pm$1.3}} & \makecell{23.0 \\ {\footnotesize $\pm$0.6}} &
  \makecell{69.8 \\ {\footnotesize $\pm$2.4}} & \makecell{95.6 \\ {\footnotesize $\pm$0.4}} & \makecell{93.2 \\ {\footnotesize $\pm$0.1}} &
  \makecell{38.0 \\ {\footnotesize $\pm$1.4}} & \makecell{73.2 \\ {\footnotesize $\pm$0.2}} & \makecell{65.6 \\ {\footnotesize $\pm$1.9}} \\[0.5em]
+ Misleading Stereotype & 
  \makecell{50.5 \\ {\footnotesize $\pm$0.9}} & \makecell{31.0 \\ {\footnotesize $\pm$1.5}} & \makecell{20.1 \\ {\footnotesize $\pm$2.1}} & 
  \makecell{49.0 \\ {\footnotesize $\pm$2.1}} & \makecell{28.3 \\ {\footnotesize $\pm$0.9}} & \makecell{21.2 \\ {\footnotesize $\pm$1.0}} &
  \makecell{37.9 \\ {\footnotesize $\pm$1.0}} & \makecell{73.1 \\ {\footnotesize $\pm$0.0}} & \makecell{63.5 \\ {\footnotesize $\pm$0.8}} \\[0.5em]
\midrule
Volatility (Std. Dev.) & $\pm$29.1 & $\pm$33.5 & $\pm$36.5 & $\pm$12.4 & $\pm$28.6 & $\pm$30.9 & $\pm$2.5 & $\pm$0.5 & $\pm$1.4 \\
\bottomrule
\end{tabular}
\caption{NLI performance on the TOEFL4 Dataset across different models, experimental setups, and hint conditions. Values represent macro-averaged F1 scores. Results shown with a standard deviation (mean $\pm$ SD) are averaged over the experimental runs. The final row, Volatility (Std.\ Dev.), reports the standard deviation of the mean F1 scores across the five evaluation tasks as a measure of performance consistency; lower values indicate better robustness to the potentially misleading augmentations. L3-70B: Llama-3.3-70B; L3-8B: Llama-3.1-8B; Kimi-K2: Kimi-K2-Instruct.}
\label{tab:nli_results}
\end{table*}

\paragraph{Syntax Expert} 
This agent focuses exclusively on identifying and classifying grammatical and structural deviations from Standard English. 
Its analysis includes subject-verb agreement errors, non-standard word order (e.g., modifier placement, verb positioning), issues in clause construction, and incorrect use of grammatical function words (articles, prepositions) related to syntactic rules.
See Figure \ref{fig:syntax-prompt} for the full prompt.

\paragraph{Lexical Expert} 
The role of this agent is to scrutinize lexical phenomena. 
Its scope includes orthographic errors (misspellings), morphological errors (incorrect word forms), inappropriate word choices (lexical selection), non-standard collocations (word pairings), potential false cognates (e.g., \textit{sensible} in place of \textit{sensitive} due to Italian \textit{sensibile}), and malapropisms (\textit{illicit} vs. \textit{elicit}).
See Figure \ref{fig:lexical-prompt} for the full prompt.

\paragraph{Idiomatic Language and Translation Expert} 
This agent specializes in analyzing the use of multi-word expressions, idioms, metaphors, and figurative language. It identifies odd phrasing, potential literal translations of L1 idioms (calques), and other misuses of standard English idiomatic or figurative expressions, focusing on deviations in non-literal language.
See Figure \ref{fig:idiomatic-prompt} for the full prompt.

\paragraph{Language Identification Expert}
This component serves as the decision maker and is the only expert in our agentic approach that is explicitly aware of the final goal: identifying the native language (L1) of the author. 
Crucially, this expert does not have direct access to the original input text. 
Instead, it is tasked with synthesizing the analyses provided by the other specialized experts. 
Based on these abstracted findings and its internal knowledge of L1 interference patterns, the investigator makes the final NLI prediction. 
This constraint ensures that the NLI decision is based only on the categorized linguistic features identified by the experts, and reduces the risk of it being able to shortcut the analysis based on surface-level clues in the original text.
The prompt for this expert is provided in Figure \ref{fig:agent-prompt}.

\section{Results}
\label{sec:results}

The main performance results on the TOEFL4 dataset are presented in Table~\ref{tab:nli_results}. Detailed results for the W\&I dataset, which exhibit similar trends, are shown in Appendix~\ref{app:wni_results} (Table~\ref{app:nli_results_wni}).

\paragraph{\textbf{How do superficial clues affect model performance under direct prompting?}}
Comparing the macro-averaged F1 scores achieved by the direct-prompting baseline, we can see substantial differences across the five experimental settings.
With no added bias signature (i.e., unmodified input text) the larger models (Llama-3.3-70B and Kimi-K2) achieve almost perfect accuracy ($\approx$94--97\%). 
Even so, adding supportive signatures resolves the few remaining classification errors and pushes performance to $\approx$100\% macro-F1. 
The effect is most pronounced for Llama-3.1-8B, which improves by about 25 points with a supportive signature ($72.7\to 97.8$). 
A similar pattern holds on W\&I (Table~\ref{app:nli_results_wni}): supportive hints yield $\approx$10-point gains for Llama-3.3-70B and Kimi-K2, with even larger improvements for Llama-3.1-8B.

\paragraph{\textbf{What is the influence of misleading information?}}
Conversely, the occurrence of misleading signatures drastically degrade performance for the direct prompt baseline. 
On both datasets, applying this augmentation to the input texts results in near or below chance-level performance for all models, with the notable exception of Llama-3.1-8B, which maintains a higher level of accuracy under the misleading stereotype setting.
One potential hypothesis for this is that due to the model's smaller capacity, it is less susceptible to influences relating to cultural and linguistic stereotypes, which aligns with previous work showing that larger models encode these relationships better than their smaller counterparts \citep{goerge_etal_2025}.
Importantly for our task, after inspecting the models' reasoning output associated with its predictions, we observe that models consistently fail to acknowledge the bias introduced by the signature.
Instead, their outputs are often characterized by hallucinations (e.g., fabricating linguistic rules that do not exist) and generating explanations that are irrelevant to the actual text \cite{hicks2024chatgpt}, in an attempt to rationalize a decision that can primarily be attributed to those superficial clues and shortcuts.

\paragraph{\textbf{Can information redaction mitigate these shortcuts?}}
Looking at the performance of the direct prompting approach with our redacted preprocessing, we can see that masking named entities allows us to maintain comparable performance with no modifications, supportive hints, and the misleading learner signature. 
This makes sense since the redaction step effectively neutralizes the bias introduced by the learner signatures.
However, results for the stereotype setting indicate that redaction is ineffective against stereotype-based hints since these typically do not contain relevant named entities.
This highlights the shortcomings of such a preprocessing step as a strategy to achieve greater consistency in the NLI task.

\begin{figure*}[htbp!]
    \centering
    \hspace*{1.7cm} 
    \includegraphics[width=0.85\textwidth]{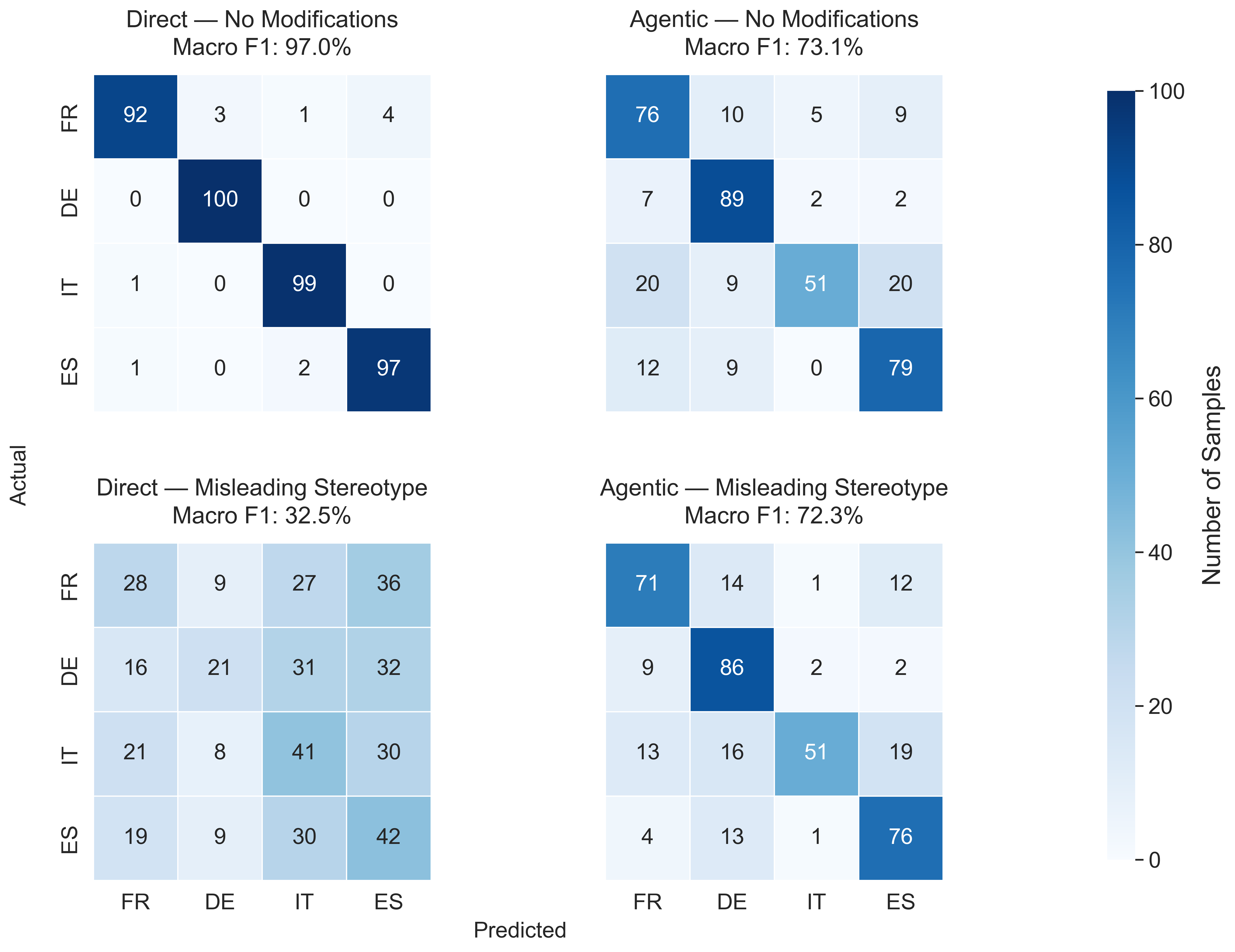}
    \caption{
        Confusion matrices showing Llama-3.3-70B's performance on the TOEFL4 dataset for the direct prompting strategy (left) and the agentic strategy (right).
        Macro-F1 scores reported here are from a single run and thus differ slightly from Table \ref{tab:nli_results}, which provides aggregate values over multiple runs.
        }
    \label{fig:comparison_cm}
\end{figure*}

\paragraph{\textbf{How does the agentic approach perform under the experimental conditions?}}
For our proposed agentic approach, we observe a clear drop in peak accuracy compared to the two baseline approaches, with Llama-3.3-70B achieving only 73.7\% on the unmodified input text.
Importantly, however, when comparing this performance across all experimental conditions, we observe far greater consistency compared to the results attained from direct prompting strategies, as reflected by the low on-aggregate volatility scores (standard deviation across conditions). Overall, this agentic workflow consistently trades off peak accuracy with markedly higher stability and robustness to adversarial input perturbations. One potential explanation for the drop in performance is that this workflow relies on upstream expert analysis, especially grammatical error detection, before the final synthesis. In cases where texts contain few irregularities, only limited informative fragments can be provided to the decision maker, making the final prediction considerably more challenging.
Furthermore, the benchmark performance in the unmodified setting may be an overestimate of true L1 identification capabilities, as texts within benchmarks such as TOEFL4 often contain supportive stereotypical names or cultural cues that models can leverage as shortcuts. Therefore while the agentic approach can decrease the overall performance relative to direct prompting, we argue that by focusing on grammatical evidence instead of spurious cues, it provides a more realistic view on LLMs' NLI capabilities. 

\section{Further Analysis}
\label{sec:further_analysis}

\begin{table}[htbp!]
\centering
\begin{tabular}{@{}lc@{}}
\textbf{Feature Combination} & \textbf{F1} \\
\midrule
\multicolumn{2}{@{}l@{}}{\textit{All Experts}} \\
\quad Syntax + Lexical + Idiomatic & 73.8 \\
\addlinespace[0.5em]
\multicolumn{2}{@{}l@{}}{\textit{Individual Experts}} \\
\quad Syntax only & 62.8 \\
\quad Lexical only & 55.5 \\
\quad Idiomatic only & 70.9 \\
\bottomrule
\end{tabular}
\caption{
Ablation study showing the influence of individual experts in the agentic classification configuration on the TOEFL4 dataset. Scores report macro-averaged F1 score. 
}
\label{tab:ablation_study}
\end{table}

\paragraph{\textbf{Which agent components are most informative?}}
In order to quantify the standalone information each analysis contributes relative to the full system, we perform a leave-one-in ablation that relies on only a single expert at a time.
On TOEFL4, the full configuration (“All Experts”) attains 73.7 macro-F1, while single-expert runs achieve 62.8 (Syntax), 55.5 (Lexical), and 70.9 (Idiomatic) macro-F1 (Table~\ref{tab:ablation_study}). 
These results indicate that all components capture some useful signals, however, the idiomatic expert performs best among single components. 
One possible explanation is that the features identified by the idiomatic expert can also reveal other grammatical and lexical clues, which overlap with features extracted by other experts. 
For example, direct translations from L1, flagged by the idiomatic expert, often introduce word-order irregularities that would also be captured by the syntactic expert. On the other hand, this may be due to a lack of constraints specified in the idiomatic expert prompt itself (Figure~\ref{fig:idiomatic-prompt}).
We suspect that refinements to these expert prompts could result in stricter disentanglement, trading off standalone performance for clearer complementarity among experts.
Overall, however, from this ablation, we can see that the combination of features extracted by multiple experts results in best performance.

\paragraph{Error Analysis}

Figure \ref{fig:comparison_cm} presents confusion matrices analyzing how prediction errors differ between the direct prompting strategy and our proposed agentic approach. In the "No Modifications" condition, incorrect predictions primarily occur between related languages (Italian, French, and Spanish). For the direct prompting strategy (left), this confusion is minimal, in line with its high macro-F1 score. In contrast, the agentic approach (right) exhibits more pronounced errors, particularly struggling to classify Italian texts, which are frequently misidentified as Spanish or French. While this performance decrease is undesirable, we conjecture that it more accurately reflects the challenging nature of the NLI task, especially when differentiating between the native L1 of closely related languages. In the "Misleading Stereotype" condition, the agentic approach maintains its accuracy, whereas the direct method's performance collapses to slightly above the random chance baseline, even confusing predictions across different language families.

\section{Discussion}
\label{discussion}
Our findings offer several insights into LLM behavior on NLI tasks and the potential of structured approaches to enhance robustness.

\paragraph{\textbf{Why does high benchmark performance not equate to task performance?}}
While LLMs have been shown to achieve near-perfect NLI accuracy on benchmarks, leading to speculation about data leaks \citep{goswami2025multilingual}, our evaluation on the recent W\&I dataset (released post-model training) suggests an alternative explanation: this performance stems from reliance on superficial clues rather than linguistic analysis. We observed that model predictions are significantly influenced by the occurrence of culturally indicative features, which can also act as prevalent supportive hints in many benchmark texts based on learner corpora.

\paragraph{\textbf{How effective are simple mitigation strategies against superficial clues?}}
Our results indicate that simple mitigation strategies are largely ineffective. Prompt-based instructions to disregard superficial clues failed to prevent models from being influenced by them. 
Similarly, the redaction of named entities, while removing some obvious hints, proved insufficient. 
This approach not only fails to address non-entity-based false clues, but it also risks eliminating genuine linguistic evidence such as L1-influenced misspellings in redacted words themselves. 
Extending a redaction-based approach beyond clearly defined named entities with the aim also capturing implicit clues such as our misleading stereotype signatures would additionally result in a high degree of information loss, which limits its practical applicability for real-world forensic texts, where preserving as much linguistic signal as possible is paramount.

\paragraph{\textbf{What are the implications and future directions for NLI?}}
The skepticism of courts towards uninterpretable computational evidence is well-documented, with judges rightly questioning methodologies where the reasoning behind a conclusion cannot be scrutinized \citep{grant2022idea}. Our proposed agentic approach, by emulating task decomposition, presents a promising, though more computationally intensive, direction for developing NLI systems that are more resistant to superficial biases. 
The key advantage lies in promoting a systematic, evidence-driven analysis that demonstrates LLM over-reliance on easily exploitable signals. 
Future work should focus on optimizing this workflow by refining agent interactions, developing more sophisticated evidence synthesis mechanisms for the coordinator. 
For example, agents could be equipped with tools to consult external knowledge bases, allowing them to base their analysis on verifiable evidence rather than just their internal knowledge. 
In parallel, methods to dynamically weight and prioritize each agent's contribution could be explored to optimize the final synthesis. 
Improving performance without sacrificing self-consistency remains a central goal for reliable AI in sensitive domains such as forensic sciences.

\section{Conclusion}
In this work, we investigated the tendency of LLMs to rely on superficial clues and take shortcuts in the NLI task, rather than engaging with the underlying linguistic patterns indicative of L1. We introduced adversarial hints, encompassing both explicit L1 learner signatures and stereotype statements, into benchmark texts to probe this behavior. Our findings demonstrate that LLMs are significantly influenced by such salient, yet potentially misleading, information, even when explicitly instructed to disregard it. Simple mitigation strategies, including direct prompt-based instructions or named entity redaction, proved insufficient to consistently prevent models from prioritizing these superficial signals. As a more robust alternative, we proposed and evaluated a decomposed agentic pipeline. This approach assigns specialist agents to analyze distinct sets of linguistic features, and a central coordinator agent to synthesize detailed findings for the final NLI prediction. This structured methodology yielded more consistent and robust performance on two datasets of English learner texts. By forcing decisions to be grounded in specific, itemized linguistic evidence rather than holistic, potentially biased impressions, the agentic approach offers a more structured and robust process.

Our results underscore the significant challenges in ensuring that LLMs adhere to nuanced instructions and mitigate biases stemming from either explicit or implicit clues. The proposed agentic workflow, by emulating a decomposed expert analysis, represents a promising direction for developing more consistent and bias-resistant LLM applications in sensitive domains such as forensic linguistics. Future work could focus on refining inter-agent communication protocols, enhancing the granularity of linguistic feature analysis within specialist agents, and exploring methods for dynamically weighting evidence from different linguistic experts.

\section*{Limitations}
\label{sec:limitations}

While our proposed agentic pipeline demonstrates significant improvements in robustness for NLI as compared to the baselines, this study has several limitations that offer avenues for future research:

\paragraph{Scope of adversarial experiments.} Our investigation into the influence of misleading clues primarily focused on the impact of relatively salient, content-based features, such as appended learner signatures (names, locations) and explicit stereotype statements. The broader field of authorship obfuscation also considers more sophisticated adversarial attacks where LLMs or malicious actors might actively attempt to impersonate specific linguistic features to convincingly mimic a target L1 background \cite{alperin2025masks}. Developing defenses against such advanced linguistic impersonation remains a critical area for future work.

\paragraph{Dataset representativeness and low-resource scenarios.} Our experiments were conducted using publicly available L2 English learner corpora. While standard for NLI research due to reliable meta-information, these datasets may not fully represent the diversity and constraints of real-world scenarios, which can include texts varying greatly in domain, style, and length, often constituting low-resource settings with only a few sentences per author. Future work should evaluate and adapt our approach to these more challenging conditions.

\paragraph{Cross-linguistic generalizability.} This study concentrated exclusively on English as L2. The specific linguistic interference patterns and the efficacy of the agentic decomposition might differ for other L1-L2 pairings. Exploring the adaptability and performance of this agentic NLI approach across a wider range of source and target languages is left for future work.

\section*{Ethical Considerations}
Our research exclusively utilized publicly available L2 English learner corpora: the pseudonymous W\&I corpus \citep{nicholls2024write} and TOEFL11 \citep{blanchard2013toefl11}, which contains no personally identifiable information. We acknowledge the broad societal implications of authorship analysis, including potential risks to the security and privacy of individuals \cite{saxena2025responsible}. Therefore, our agentic pipeline is presented strictly for research purposes within controlled settings, primarily to study the impact of bias in existing AI systems and explore methods for enhancing robustness. This work is not intended for deployment in critical real-world applications. As detailed in our Limitations (Section \ref{sec:limitations}), we also recognize that our efforts to mitigate bias are not exhaustive, and further research is needed.

\section*{Acknowledgments}
This work was supported by the collaboration between the University of Zurich and PRODAFT as part of the Innosuisse innovation project 103.188 IP-ICT conducted at Linguistic Research Infrastructure (LIRi).

\bibliography{custom}

\clearpage 

\appendix

\section{Experimental Details and Model Parameters}
\label{appendix_experiments}

\subsection*{API Services and Client}
For querying LLMs, we used the Groq API.\footnote{\url{https://console.groq.com/docs/api}}  Groq is a LLM service provider that does not retain or train on user data sent through its API.\footnote{\url{https://groq.com/privacy-policy/}}
This ensures that data from the research corpora was not leaked to the service providers, aligning with the dataset's licensing conditions and responsible NLP guidelines.

\subsection*{Data Artifacts}
\label{sec:dataset_compliance}
The datasets for the task were sourced from two established learner corpora:
\begin{itemize}
    \item The TOEFL11 Corpus \citep{blanchard2013toefl11}, obtained under license from the Linguistic Data Consortium (LDC).
    \item The Write \& Improve Corpus 2024 \citep{nicholls2024write}, obtained under a research-use-only license from Cambridge University Press \& Assessment.
\end{itemize}
Our use of both datasets strictly adhered to their respective licensing terms, which permit non-commercial research and educational purposes. 

\subsection*{Compute Budget}
We estimate the total compute budget based on the API usage on Groq API to be approximately 150 Euro. 


\onecolumn

\section{Task Hints}
\label{app:signatures}

\begin{table*}[htbp!]
\centering
\begin{tabular}{p{2cm}p{6cm}p{0.8cm}p{4cm}}
\textbf{L1} & \textbf{Stereotype Statement} & & \textbf{Learner Signature} \\
\midrule
Italian & 
A fun fact about me: I usually start my day with a quick espresso standing at the bar. And please, no cappuccinos in the afternoon! & &
\begin{minipage}[t]{5.5cm}
Best regards,\\
Giulia Rossi\\
Milan English Academy\\
Via Monte Napoleone 18\\
20121 Milan, Italy
\end{minipage} \\
\midrule
Spanish & 
A fun fact about me: A short nap after lunch is essential for me. My culture even has a word for it: siesta. & &
\begin{minipage}[t]{5.5cm}
Best regards,\\
María García\\
Madrid Language School\\
Calle de Alcalá 45\\
28014 Madrid, Spain
\end{minipage} \\
\midrule

German & 
A fun fact about me: I love using bottle-return machines and driving on roads with no speed limit. &  &
\begin{minipage}[t]{5.5cm}
Best regards,\\
Johannes Müller\\
Berlin English Institute\\
Kurfürstendamm 123\\
10711 Berlin, Germany
\end{minipage} \\
\midrule
French & 
A fun fact about me: Where I'm from, there's nothing a fresh baguette and a good protest can't fix! & &
\begin{minipage}[t]{5.5cm}
Best regards,\\
Lucas Dubois\\
Paris Language Center\\
10 Rue de Rivoli\\
75001 Paris, France
\end{minipage} \\
\bottomrule
\end{tabular}
\caption{Custom L1 signatures used to augment inputs texts as supportive and misleading hints in our experiments.}
\label{tab:signatures}
\end{table*}

\clearpage 
\section{LLM Prompts} 

\footnotesize
\label{app:prompts_sequential} 

\begin{figure*}[h!]
\centering
\begin{minipage}{\textwidth}
\begin{lstlisting}[style=classicprompt]
You are a forensic linguistics expert that reads texts written by non-native authors to identify their native language.
Use clues such as spelling errors, word choice, syntactic patterns, and grammatical errors to decide.
Disregard any contextual information, such as names, addresses, institutions, locations, or cultural references in the text.
Analyze the input and identify the native language of the author as one of the following: French, Spanish, Italian, German. 

Text: {text}
\end{lstlisting}
\end{minipage}
\vspace{-0.5cm}
\caption{Single-shot Forensic Linguist Prompt for Baseline 1 and 2.}
\label{fig:base-prompt}
\end{figure*}

\begin{figure*}[h!]
\centering
\begin{minipage}{\textwidth}
\begin{lstlisting}[style=classicprompt]
You are a language expert. Your task is to analyze the following L2 English text exclusively for syntactic errors. 
    The other experts already cover lexical and idiomatic errors on the word level. 
    Focus on grammatical rules like word order, subject-verb agreement, clause structure, tense usage, and modifier placement.,

For each syntactic error identified, include:
1. The `error_type` (e.g., "Incorrect word order", "Subject-verb disagreement").
2. A minimal `explanation` of the grammatical problem.
3. The specific `phrase` (e.g., 3-5 words) where the error occurs.

Do not provide any cultural analysis and references in your error explanations.
Return the output as a JSON array. If no syntactic errors are found, return an empty array.

Text: {text}
\end{lstlisting}
\end{minipage}
\vspace{-0.5cm}
\caption{Syntax expert prompt used in our agentic method.}
\label{fig:syntax-prompt}
\end{figure*}

\begin{figure*}[h!]
\centering
\begin{minipage}{\textwidth}
\begin{lstlisting}[style=classicprompt]
You are a language expert. Your task is to analyze the following L2 English text exclusively for lexical errors.

Focus on identifying and explaining lexical errors where a word is:
- Spelled incorrectly (e.g., false cognates such as "adressse" instead of "address")
- A malapropism (e.g., "illicit" instead of "elicit")
- A false cognate (e.g., "sensible" instead of "sensitive")

For each error, include the `word` containing the lexical error, the `error_type`, and a minimal `explanation`.
Do not provide any cultural analysis and references in your error explanations.
Return the output as a JSON array. If no lexical errors are found, return an empty array.

Text: {text}
\end{lstlisting}
\end{minipage}
\vspace{-0.5cm}
\caption{Lexical expert prompt used in our agentic method.}
\label{fig:lexical-prompt}
\end{figure*}

\begin{figure*}[h!]
\centering
\begin{minipage}{\textwidth}
\begin{lstlisting}[style=classicprompt]
You are a language expert. Your task is to analyze the following L2 English text exclusively for idiomatic errors.
Focus on identifying incorrect, awkward, or misused multi-word expressions and figurative expressions. 
These are typically phrases where the overall meaning is not deducible from the literal meanings of the individual words. Pay attention to:
    - Potential mistranslations or literal translations.
    - Violations of common idiomatic expressions in standard English (e.g., "heavy rain" instead of "strong rain").

For each error, include the problematic `expression`, the `error_type`, and a minimal `explanation` of why it's an idiomatic error.
Do not provide any cultural analysis and references in your error explanations.
Return the output as a JSON array. If no idiomatic errors are found, return an empty array.

Text: {text}
\end{lstlisting}
\end{minipage}
\vspace{-0.5cm}
\caption{Idiom expert prompt used in our agentic method.}
\label{fig:idiomatic-prompt}
\end{figure*}

\begin{figure*}[ht!]
\centering
\begin{minipage}{\textwidth}
\begin{lstlisting}[style=classicprompt]
You are a forensic linguistics expert that reads texts written by non-native authors to identify their native language.
You will be a given an expert analysis of the text. Use clues such as spelling errors, word choice, and grammatical errors to decide.

{analysis}

Disregard any contextual information, such as names, addresses, institutions, locations, or cultural references in the text.
Analyze the input and identify the native language of the author as one of the following: French, Spanish, Italian, German. 
Provide your analysis in the JSON format.
Text:
{text}
\end{lstlisting}
\end{minipage}
\vspace{-0.5cm}
\caption{Final forensic linguistic expert prompt used in our agentic method.}
\label{fig:agent-prompt}
\end{figure*}

\clearpage

\section{The Results on the Write \& Improve Benchmark}
\label{app:wni_results}
\begin{table*}[ht]
\centering
\setlength{\tabcolsep}{4pt}
\renewcommand{\arraystretch}{1.4}
\sisetup{table-format=1.2}
\begin{tabular}{@{}l c c c c c c c c c@{}}
& \multicolumn{3}{c}{\textbf{Direct}} & \multicolumn{3}{c}{\textbf{Redacted}} & \multicolumn{3}{c}{\textbf{Agentic}} \\
\cmidrule(lr){2-4} \cmidrule(lr){5-7} \cmidrule(lr){8-10}
\textbf{Experimental Setting} & 
  {\rot{L3-8B}} & {\rot{L3-70B}} & {\rot{Kimi-K2}} &
  {\rot{L3-8B}} & {\rot{L3-70B}} & {\rot{Kimi-K2}} &
  {\rot{L3-8B}} & {\rot{L3-70B}} & {\rot{Kimi-K2}} \\
\midrule
No modification & 
  \makecell{59.6 \\ {\footnotesize $\pm$0.3}} & \makecell{89.9 \\ {\footnotesize $\pm$0.8}} & \makecell{90.8 \\ {\footnotesize $\pm$1.1}} &
  \makecell{59.3 \\ {\footnotesize $\pm$2.1}} & \makecell{91.3 \\ {\footnotesize $\pm$0.5}} & \makecell{89.4 \\ {\footnotesize $\pm$1.2}} &
  \makecell{32.1 \\ {\footnotesize $\pm$1.3}} & \makecell{62.2 \\ {\footnotesize $\pm$0.2}} & \makecell{56.4 \\ {\footnotesize $\pm$1.9}} \\[0.5em]
+ Supportive Signature & 
  \makecell{97.2 \\ {\footnotesize $\pm$0.7}} & \makecell{99.4 \\ {\footnotesize $\pm$0.1}} & \makecell{100.0 \\ {\footnotesize $\pm$0.0}} &
  \makecell{58.8 \\ {\footnotesize $\pm$1.1}} & \makecell{89.1 \\ {\footnotesize $\pm$0.2}} & \makecell{90.4 \\ {\footnotesize $\pm$1.6}} &
  \makecell{39.4 \\ {\footnotesize $\pm$2.1}} & \makecell{65.6 \\ {\footnotesize $\pm$0.1}} & \makecell{57.7 \\ {\footnotesize $\pm$0.5}} \\[0.5em]
+ Supportive Stereotype & 
  \makecell{83.6 \\ {\footnotesize $\pm$1.7}} & \makecell{99.9 \\ {\footnotesize $\pm$0.1}} & \makecell{99.6 \\ {\footnotesize $\pm$0.1}} &
  \makecell{86.8 \\ {\footnotesize $\pm$0.5}} & \makecell{100.0 \\ {\footnotesize $\pm$0.0}} & \makecell{99.4 \\ {\footnotesize $\pm$0.1}} &
  \makecell{39.7 \\ {\footnotesize $\pm$1.9}} & \makecell{71.6 \\ {\footnotesize $\pm$0.5}} & \makecell{57.2 \\ {\footnotesize $\pm$1.1}} \\[0.5em]
+ Misleading Signature & 
  \makecell{12.2 \\ {\footnotesize $\pm$2.1}} & \makecell{18.8 \\ {\footnotesize $\pm$1.5}} & \makecell{15.2 \\ {\footnotesize $\pm$2.0}} &
  \makecell{59.1 \\ {\footnotesize $\pm$1.3}} & \makecell{88.7 \\ {\footnotesize $\pm$0.0}} & \makecell{90.3 \\ {\footnotesize $\pm$0.6}} &
  \makecell{32.1 \\ {\footnotesize $\pm$0.2}} & \makecell{62.6 \\ {\footnotesize $\pm$1.2}} & \makecell{54.3 \\ {\footnotesize $\pm$1.3}} \\[0.5em]
+ Misleading Stereotype & 
  \makecell{37.9 \\ {\footnotesize $\pm$1.5}} & \makecell{23.6 \\ {\footnotesize $\pm$1.4}} & \makecell{11.9\\ {\footnotesize $\pm$0.2}} & 
  \makecell{35.1 \\ {\footnotesize $\pm$0.1}} & \makecell{18.3 \\ {\footnotesize $\pm$1.2}} & \makecell{11.3 \\ {\footnotesize $\pm$0.3}} &
  \makecell{29.7 \\ {\footnotesize $\pm$1.5}} & \makecell{58.3 \\ {\footnotesize $\pm$1.3}} & \makecell{52.1 \\ {\footnotesize $\pm$3.7}} \\[0.5em]
\midrule
Volatility (Std. Dev.) & $\pm$19.3 & $\pm$35.3 & $\pm$40.4 & $\pm$16.4 & $\pm$32.3 & $\pm$32.3 & $\pm$4.1 & $\pm$4.4 & $\pm$2.1 \\
\bottomrule
\end{tabular}
\caption{Macro-averaged F1 scores for NLI performance on the W\&I Dataset across different models, experimental setups, and hint conditions. Results shown with a standard deviation (mean $\pm$ SD) are averaged over three independent runs. The final row, Volatility (Std.\ Dev.), reports the standard deviation of the mean F1 scores across the five evaluation tasks as a measure of performance consistency; lower values indicate better robustness to the potentially misleading augmentations. 
L3-8B: Llama-3.1-8B; L3-70B: Llama-3.3-70B; Kimi-K2: Kimi-K2.
}
\label{app:nli_results_wni}
\end{table*}

\end{document}